\pdfoutput=1

\documentclass[11pt]{article}

\usepackage[final]{latex/acl}

\usepackage{times}
\usepackage{latexsym}

\usepackage[T1]{fontenc}

\usepackage[utf8]{inputenc}

\usepackage{microtype}

\usepackage{inconsolata}

\usepackage{graphicx}
\usepackage{booktabs}
\usepackage{xcolor}
\usepackage{colortbl}

\usepackage{mdframed}
\usepackage{amsmath}
\usepackage{multirow}

\title{XLQA: A Benchmark for Locale-Aware Multilingual Open-Domain Question Answering}

\author{
 \textbf{Keon-Woo Roh\textsuperscript{1}},
 \textbf{Yeong-Joon Ju\textsuperscript{1}},
 \textbf{Seong-Whan Lee\textsuperscript{1}}
\\
 \textsuperscript{1}Department of Artificial Intelligence, Korea University
\\
 \texttt{\{ro\_keonwoo, yj\_ju, sw.lee\}@korea.ac.kr}
}

\begin{document}
\maketitle
\begin{abstract}

Large Language Models (LLMs) have shown significant progress in Open-domain question answering (ODQA), yet most evaluations focus on English and assume locale-invariant answers across languages. This assumption neglects the cultural and regional variations that affect question understanding and answer, leading to biased evaluation in multilingual benchmarks. To address these limitations, we introduce XLQA, a novel benchmark explicitly designed for locale-sensitive multilingual ODQA. XLQA contains 3,000 English seed questions expanded to eight languages, with careful filtering for semantic consistency and human-verified annotations distinguishing locale-invariant and locale-sensitive cases. Our evaluation of five state-of-the-art multilingual LLMs reveals notable failures on locale-sensitive questions, exposing gaps between English and other languages due to a lack of locale-grounding knowledge. We provide a systematic framework and scalable methodology for assessing multilingual QA under diverse cultural contexts, offering a critical resource to advance the real-world applicability of multilingual ODQA systems. Our findings suggest that disparities in training data distribution contribute to differences in both linguistic competence and locale-awareness across models.

\end{abstract}
\section{Introduction}

\begin{figure}
\centering
\includegraphics[width = 1.\columnwidth]{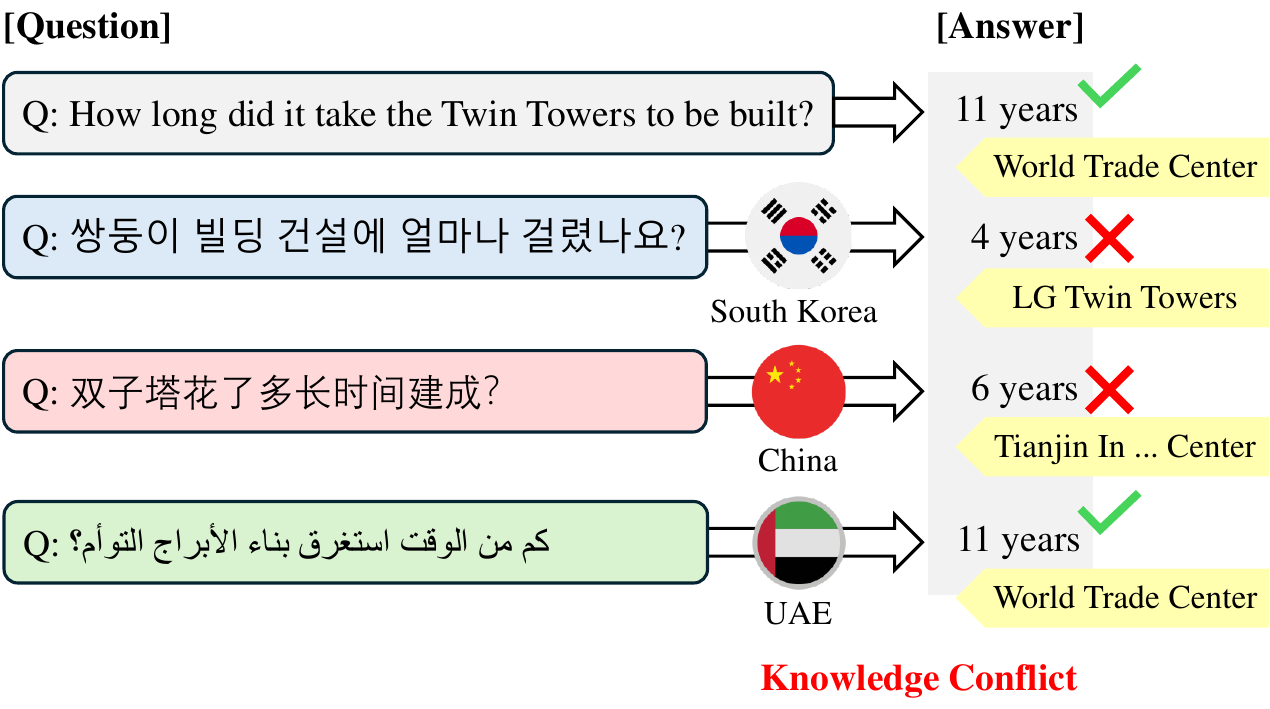}
\captionof{figure}{
    \textbf{Knowledge conflict in multilingual ODQA.} Although all versions of the question aim to ask how long it took to build the "Twin Towers", different languages elicit different answers based on locale-variant understanding. While English and Arabic refer to the World Trade Center (11 years), Korean and Chinese interpret "Twin Towers" as the LG Twin Towers and Tianjin IFC, respectively.
}
\vspace{-8pt}
\label{fig:ex}
\end{figure}

Open-domain question answering (ODQA) aims to generate accurate and natural language answers to user queries without explicit domain constraints or provided context \citep{chen-etal-2017-reading, karpukhin-etal-2020-dense}. Recently, large language models (LLMs) \citep{NEURIPS2020_1457c0d6, anil2312gemini, workshop2022bloom} have driven significant advances in ODQA by generating correct and natural answers. Despite strong advances in ODQA, most efforts have focused on English, leaving multilingual capabilities that remain relatively under-explored. This gap underscores the need for multilingual ODQA benchmarks that assess performance across languages~\citep{maxutov-etal-2024-llms}.

To evaluate multilingual ODQA systems, existing benchmarks, such as MLQA \citep{lewis-etal-2020-mlqa}, MKQA \citep{longpre-etal-2021-mkqa}, and TyDiQA \citep{clark-etal-2020-tydi}, are typically constructed by translating or aligning parallel questions across multiple languages. 
These benchmarks have the locale-agnostic assumption that both the meaning of a question and its correct answer remain constant across linguistic boundaries. However, this assumption overlooks variations in meaning that arise naturally from distinct cultural or regional contexts \citep{lin-etal-2021-cultureqa, liu-etal-2024-multilingual, zhang-etal-2023-dont}. This issue introduces evaluation bias \citep{talat-etal-2022-reap} by penalizing responses that are correct within specific regional or cultural contexts. For instance, as illustrated in Fig.~\ref{fig:ex}, the answer to the question “How long did it take the Twin Towers to be built?” differs depending on which entity the question refers to: the World Trade Center in the U.S. or the LG Twin Towers in South Korea. Multilingual question requires the locale-variant references arise from differing cultural contexts and background knowledge, not merely generating translated answer. In addition, relying on naive translation to construct multilingual benchmarks risks semantic drift, where subtle shifts in meaning occur due to inadequate contextual grounding \citep{yu2023large}. While human annotation can mitigate the drift, it is costly, labor-intensive, and difficult to scale across many languages and cultures \citep{PANDEY2022102772}.

To address these challenges, we propose \textbf{XLQA}, a benchmark explicitly constructed to evaluate multilingual ODQA systems under locale-sensitive conditions. XLQA consists of 3,000 seed questions in English, each paired with a reference answer and language-specific supporting evidence. These questions are extended to eight languages, resulting in 24,000 high-quality evaluation items. We design XLQA to assess whether multilingual ODQA systems can handle locale-sensitive variation by explicitly distinguishing between two types of questions: those whose correct answers remain consistent across languages (locale-invariant), and those whose answers vary depending on regional or linguistic context (locale-sensitive).

To construct this benchmark at scale, we apply a back-translation-based filtering method to identify and remove translations that exhibit potential semantic inconsistencies. Then, we generate locale-aware answers for each semantically consistent multilingual question by producing responses based on language-specific evidence curated for each locale with an LLM. These generated answers that semantically differ from the original English answer is categorized as a potentially locale-sensitive question. Human annotators examine each candidate instance to verify the answer’s correctness and the relevance of the supporting evidence. This approach enables scalable multilingual QA dataset creation with limited human involvement, ensuring quality through selective verification rather than full manual annotation.

To demonstrate the effectiveness of this pipeline, we evaluate five multilingual LLMs on our benchmark, such as GPT-4.1 \citep{achiam2023gpt}, Qwen-3 \citep{zheng2025empirical}, Gemma-3 \citep{team2025gemma}, LLaMA-3.1 \citep{grattafiori2024llama}, and Exaone \citep{research2024exaone} under standard evaluation metrics, including exact match and F1 score. Our analysis reveals that, despite strong zero-shot and multilingual capabilities, these models frequently fail to produce appropriate answers to locale-sensitive questions. We observe differences in both language proficiency and locale-specific knowledge across models, shaped by the distribution of language data used during training. These findings highlight the limitations of existing multilingual QA benchmarks and underscore the importance of explicitly modeling cultural context in evaluation. We summarize our contributions as follows:

\begin{itemize}
    \item We introduce the first systematic framework for evaluating locale-aware correctness in multilingual QA, directly addressing the cultural insensitivity and English-centric assumptions embedded in prior benchmarks.
    \item We propose a scalable method for identifying and validating questions whose correct answers vary across regions, producing a benchmark of 3,000 high-quality question–answer–evidence triples annotated for locale sensitivity.
    \item We provide empirical evidence that current multilingual LLMs struggle with locale-grounded question answering, revealing a critical gap in their real-world applicability.
\end{itemize}

\begin{figure*}
\centering
\includegraphics[width = 1.\textwidth]{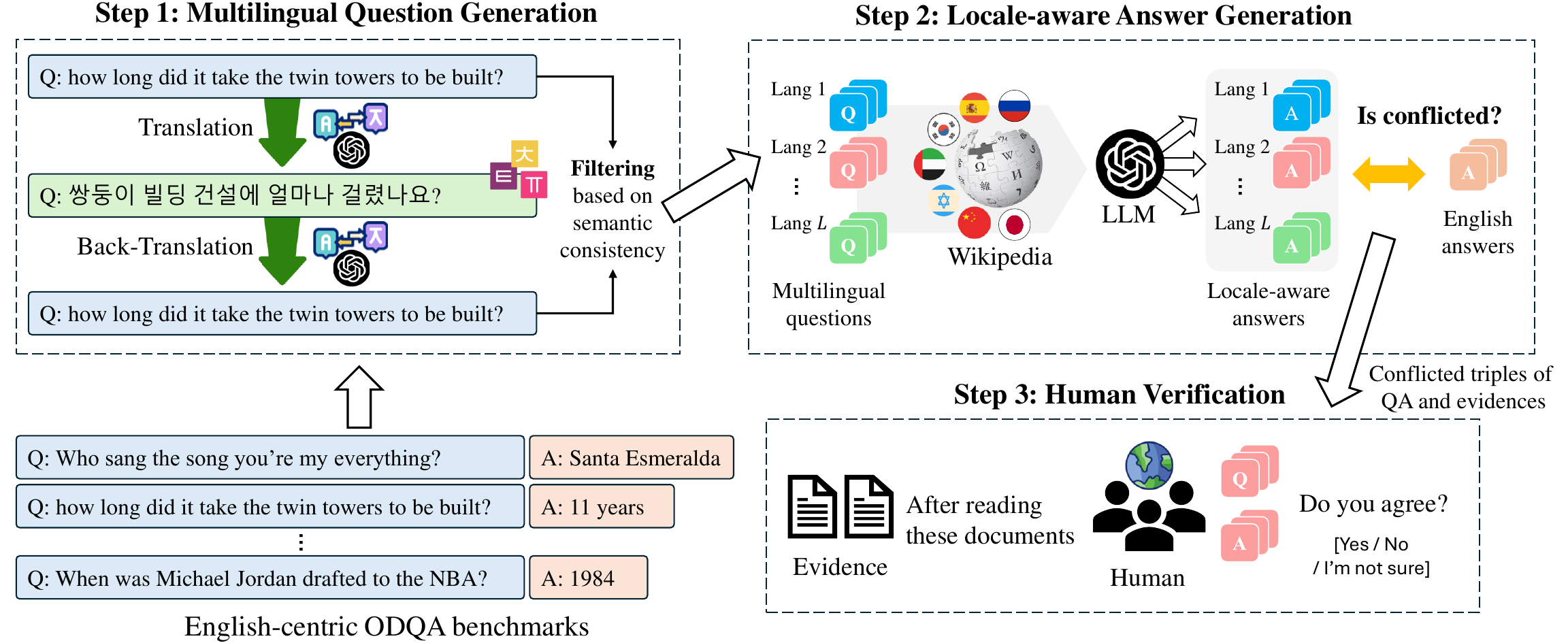}
\captionof{figure}{
     \textbf{The overall pipeline for constructing the XLQA benchmark.} The process consists of three stages: (1) \textbf{Multilingual Question Generation} generates multilingual questions based on seed questions from existing QA datasets. (2) \textbf{Locale-Aware Answer Generation} uses LLM to generate locale-aware answers. (3) \textbf{Human Verification} verifies the answers with supporting evidence. The Output is a high-quality, locale-aware multilingual QA dataset.
}
\label{fig:main}
\end{figure*}

\section{Related Works}

\subsection{Multilingual ODQA Benchmarks}

In recent years, numerous multilingual question answering (QA) benchmarks have been proposed to evaluate the performance of multilingual language models. Prominent examples include MLQA \citep{lewis-etal-2020-mlqa}, XQuAD \citep{artetxe-etal-2020-cross}, TyDiQA \citep{clark-etal-2020-tydi}, and MKQA \citep{longpre-etal-2021-mkqa}, which are widely used to compare model performance across different languages.

MLQA and XQuAD are constructed by translating English question–answer pairs into multiple target languages, and rely on the assumption that the translated versions are semantically equivalent to the original. This approach enables direct comparison across languages but may overlook subtle linguistic or cultural differences that affect answer validity. In contrast, TyDiQA enhances linguistic diversity by collecting questions written natively in each language by fluent speakers, rather than relying on translation. However, it still assumes a single ground-truth answer per question within each language, potentially limiting its ability to capture within-language ambiguity or region-specific variation. MKQA takes a different approach by sourcing questions from anonymized Google Assistant logs, reflecting more natural, real-world user queries. These questions are then manually translated into 26 languages for open-domain question answering.

While these benchmarks provide a foundation for measuring multilingual capabilities and cross-lingual consistency, they largely focus on surface-level correctness and lexical alignment. As such, they fall short of evaluating model performance in scenarios that require understanding of cultural context or locale-specific knowledge salience.

\subsection{Multilingual QA Evaluation Bias and Fairness}


Recent works \citep{global-mmlu-2024, nativqa-2024} have examined these issues from multiple perspectives. \citet{global-mmlu-2024} evaluates language models across culturally diverse multiple-choice questions. They shows that performance varies substantially across languages and regions, indicating potential cultural bias. \citet{nativqa-2024} introduces a dataset of naturally occurring, culturally aligned queries in multiple languages. Their findings highlight the limitations of translation-based benchmarks in capturing region-specific information needs.

Bias is observed in model behavior across languages with differing resource levels, particularly in the form of stereotypical associations related to gender, profession, or ethnicity. \citet{buscemi2025mind} proposes an automated evaluation framework to assess such social biases across both high- and low-resource languages. The study finds that these biases, such as associating certain professions more frequently with specific genders, tend to be more pronounced in low-resource settings, where training data is sparser and less balanced.

Similarly, \citet{saralegi-zulaika-2025-basqbbq} adapts the English-centric BBQ benchmark to Basque in order to investigate bias propagation in a typologically distant language. Their findings reveal that common bias mitigation strategies developed for English, such as data augmentation or counterfactual training, often fail to generalize effectively to underrepresented languages, underscoring the need for culturally and linguistically tailored approaches. These studies point to the need for evaluation methods that distinguish between culturally invariant and culturally dependent questions, and that reflect the diversity of real-world language use above high-resource settings.

\subsection{Evaluation for LLM-as-judges}
LLM-as-judge is a generative evaluator paradigm where LLMs are trained to produce an evaluation (natural language explanation and judgment) given the original user input, evaluation protocol (rules and criteria for evaluation), and model responses as input. JudgeLM \citep{zhu2025judgelm} formalizes this approach as a generative evaluation framework and demonstrates that LLM-based judges can approximate human evaluations in tasks such as reasoning and factual correctness. PandaLM \citep{wang2024pandalm} further investigates the reliability and robustness of LLM-based evaluators by comparing their preferences across model outputs with those of human annotators.
\section{XLQA Dataset}

To rigorously evaluate multilingual ODQA in locale-sensitive contexts, we introduce XLQA, a new benchmark constructed through our multi-stage pipeline.
This pipeline consists of three steps: multilingual question generation, locale-aware answer generation, and human verification, as illustrated in Fig.~\ref{fig:main}.

\subsection{Step 1: Multilingual Question Generation}

 We begin by collecting high-quality English seed questions from the test sets of existing ODQA benchmarks, such as MKQA~\citep{longpre-etal-2021-mkqa}, MLQA~\citep{lewis-etal-2020-mlqa}, and HotpotQA~\cite{yang-etal-2018-hotpotqa}, to ensure alignment with our evaluation objectives. To refine the seed pool, we first remove duplicate entries based on an exact match of either the question or the answer. We then filter out unanswerable questions or those lacking a reference answer, as such items prevent meaningful comparison of locale-sensitive responses. This filtering process results in the exclusion of 28.4\% of the initial seed questions.
 
For the refined seed questions, we generate multilingual questions translated into diverse target languages by utilizing GPT-4.1 (Oracle LM) \citep{achiam2023gpt}, which demonstrates strong performance in translation quality and contextual understanding. To ensure semantic consistency across the translated questions, we apply a back-translation filtering step. Each translated question is first back-translated into English. Then, the resulting back-translated version is compared against the original English question using the LLM-as-judge framework. The model is prompted to determine whether the two questions are semantically equivalent, providing a binary "yes/no" judgment. If any of the eight language translations are judged as inconsistent (i.e., the model outputs "no"), the entire question is discarded from the dataset. By discarding questions with inconsistent translations, this back-translation filtering step plays a crucial role in eliminating translation artifacts and mitigating cross-lingual meaning drift.

\subsection{Step 2: Locale-Aware Answer Generation}

To construct QA pairs that capture locale-specific variation, we generate candidate answers for the multilingual questions obtained in the previous step. For each input question, the model is prompted to generate an answer that reflects the locale associated with the language in which the question is written. For questions that are not sensitive to locale, the model is prompted to provide a general, culturally neutral answer. We leverage a retrieval-augmented generation (RAG) framework in which GPT-4.1 is connected to a web search component. This setup enables the model to generate answers grounded in verifiable external sources, providing both the response and its corresponding evidence. The retrieval process prioritizes authoritative sources, with a preference for Wikipedia. In case that relevant information is not found on Wikipedia, the system falls back to reputable news outlets.

As a post-processing step, we discard any QA pairs in which the generated reference lacks a valid URL or does not include reliable source indicators such as the keywords "wikipedia" or "news". This filtering ensures that all retained answers are grounded in verifiable and trustworthy sources.
This approach offers an efficient alternative to human annotation by enabling scalable, high-quality data generation while maintaining contextual relevance and answer verifiability.


\subsection{Step 3: Human Verification}

All candidate triples flagged for answer conflict are subjected to human verification. Annotators are provided with the question, answer, and supporting evidence for each language. They are asked to determine whether the answer is correct and supported by the evidence.
This process yields a high-quality set of QA-evidence triples, each labeled as either locale-invariant or locale-sensitive.
To ensure consistency and reduce annotation noise, we adopt a majority voting scheme across three annotators per instance. Only instances where at least two annotators agree on both correctness and sensitivity labels are retained; otherwise, the item is discarded.
Statistics on annotator agreement rates after voting are provided in Appendix Table~\ref{tab:annotation_agreement}.

\begin{table*}[t]
\centering
\resizebox{.95\textwidth}{!}{%
\begin{tabular}{lcccccccccc}
\toprule
\textbf{Lang} & \multicolumn{2}{c}{\textbf{Oracle LM}} & \multicolumn{2}{c}{\textbf{Gemma3 12B}} & \multicolumn{2}{c}{\textbf{Qwen3 14B}} & \multicolumn{2}{c}{\textbf{LLaMA3.1 8B}} & \multicolumn{2}{c}{\textbf{Exaone 7.8B}} \\
\cmidrule(r){2-3} \cmidrule(r){4-5} \cmidrule(r){6-7} \cmidrule(r){8-9} \cmidrule(r){10-11}
& EM & F1 & EM & F1 & EM & F1 & EM & F1 & EM & F1 \\
\midrule
en     & 89.11 & 90.97 & \textbf{43.26} & \textbf{52.68} & \textbf{40.73} & \textbf{49.43} & \textbf{40.38} & \textbf{50.56} & \textbf{31.44} & \textbf{39.64} \\
ar     & 87.86 & 90.05 & 18.54 & 23.62 & 11.83 & 19.30 & 8.53  & 16.92 & 3.98  & 6.04 \\
he     & 88.30 & 90.46 & 20.05 & 24.83 & 11.04 & 16.08 & 11.86 & 16.60 & 5.52  & 7.20 \\
ja     & 88.45 & 92.50 & 22.81 & 45.10 & 19.74 & 44.03 & 9.10  & 37.73 & 7.34  & 26.22 \\
ru     & 87.83 & 89.54 & 28.52 & 35.20 & 17.67 & 27.97 & 14.53 & 24.18 & 7.41  & 9.91 \\
ko     & 86.73 & 88.29 & 22.18 & 26.56 & 15.44 & 19.91 & 11.55 & 15.68 & 15.81 & 20.32 \\
zh\_cn & \textbf{89.68} & \textbf{93.41} & 16.22 & 37.91 & 26.39 & 47.57 & 11.58 & 36.48 & 7.66  & 25.22 \\
vi     & 89.39 & 91.19 & 36.55 & 44.77 & 26.83 & 39.34 & 26.45 & 38.38 & 10.95 & 14.70 \\
\midrule
Avg.    & 88.42 & 90.80 & 26.02 & 36.33 & 21.21 & 32.95 & 16.75 & 29.57 & 11.26 & 18.65 \\
\bottomrule
\end{tabular}%
}
\caption{Results of the base models on the XLQA benchmark using EM and F1 scores.}
\label{tab: main}
\end{table*}

\section{Dataset Analysis}

\subsection{Dataset Statistics}
Our benchmark consists of 3,000 question–answer–evidence triples across eight languages: English, Korean, Arabic, Hebrew, Japanese, Russian, Vietnamese, and Simplified Chinese. Each English-origin question is translated into the target languages and paired with answers and evidential support adapted to the cultural or linguistic context of the target locale.

On average, questions contain 17–40 tokens depending on language, while answers remain short (4–6 tokens). A total of 24,000 QA instances were created, including 3,000 in English and 21,000 across the seven other languages.

\subsection{Consistency Filtering Results}
To ensure semantic consistency across translations, we applied a back-translation-based filtering pipeline. 
QA pairs with substantial semantic shifts, such as changes in named entities, factual scope, or temporal modifiers, were flagged and removed. In total, 10.8\% of generated multilingual instances were discarded through this process.

We observed that the majority of filtered instances involved mistranslations of culturally specific terms or reinterpretations of ambiguous expressions that altered the intended meaning. These cases were particularly prevalent in Arabic and Hebrew, where semantic drift often resulted from incorrect rendering of proper nouns and idiomatic language. Table~\ref{tab:consistency} summarizes the number of discarded instances per language following the consistency filtering process.

\subsection{Conflict Detection}
A conflict is defined as a case where at least one language provides an answer that is semantically inconsistent with the English reference, under the assumption that such variation is due to regional knowledge or interpretation.
For each question, we collected answers across all languages and compared them using string normalization and embedding-based semantic similarity.
Questions exhibiting divergence in meaning, rather than surface expression, were manually validated as locale-sensitive. Among the 3,000 source questions, 2,356 (73.9\%) were categorized as locale-sensitive, based on the presence of conflicting answers in at least one language.
Table~\ref{tab:consistency} presents the distribution of conflicts across languages. Arabic and Hebrew displayed the highest proportion of conflicts, while Japanese and Vietnamese showed comparatively lower divergence.

\begin{table*}[t]
\centering
\resizebox{\textwidth}{!}{%
\begin{tabular}{lcccc|cccc|cccc}
\toprule
\textbf{Lang} 
& \multicolumn{4}{c|}{\textbf{\textsc{Gemma3 12B}}} 
& \multicolumn{4}{c|}{\textbf{\textsc{Qwen3 14B}}} 
& \multicolumn{4}{c}{\textbf{\textsc{Exaone 7.8B}}} \\
\cmidrule(r){2-5} \cmidrule(r){6-9} \cmidrule(r){10-13}
& \multicolumn{2}{c}{Non-Conflict} & \multicolumn{2}{c|}{Least-Conflict}
& \multicolumn{2}{c}{Non-Conflict} & \multicolumn{2}{c|}{Least-Conflict}
& \multicolumn{2}{c}{Non-Conflict} & \multicolumn{2}{c}{Least-Conflict} \\
& EM & F1 & EM & F1 & EM & F1 & EM & F1 & EM & F1 & EM & F1 \\
\midrule
\rowcolor{gray!10}en     & 59.09 & 72.25 & 37.69 & 45.79 & 64.02 & 75.36 & 32.51 & 40.28 & 53.67 & 65.28 & 23.60 & 30.59 \\
ar     & 37.06 & 46.26 & 12.01 & 15.64 & 27.80 & 40.54 & 6.20  & 11.80 & 8.90 & 12.21 & 2.25 & 3.86 \\
he     & 38.63 & 47.18 & 13.50 & 16.94 & 23.71 & 32.16 & 6.58  & 10.41 & 9.63 & 12.49 & 4.07 & 5.33 \\
ja     & 41.16 & \textbf{67.03} & 16.34 & 37.36 & 41.03 & 65.30 & 12.22 & 36.53 & 15.28 & 38.24 & 4.54 & \textbf{21.98} \\
ru     & 47.41 & 57.02 & 21.86 & 27.50 & 30.69 & 49.38 & 13.07 & 20.42 & 10.95 & 15.48 & 6.15 & 7.95 \\
ko     & 39.35 & 46.06 & 16.13 & 19.69 & 31.05 & 38.11 & 9.93  & 13.49 & \textbf{30.93} & \textbf{38.48} & \textbf{10.48} & 13.91 \\
\textnormal{zh\_cn} & 27.32 & 56.12 & 12.31 & 31.49 & \textbf{50.54} & \textbf{71.87} & 17.87 & \textbf{39.00} & 15.16 & 37.15 & 5.01 & 21.01 \\
vi     & \textbf{51.62} & 63.79 & \textbf{31.24} & \textbf{38.07} & 41.40 & 61.34 & \textbf{21.69} & 31.58 & 18.05 & 24.30 & 8.45 & 11.31 \\
\midrule
\textbf{Average} & 42.70 & 56.96 & 20.13 & 29.06 & 38.78 & 54.26 & 15.01 & 25.44 & 20.32 & 30.45 & 8.07 & 14.49 \\
\bottomrule
\end{tabular}%
}
\caption{EM and F1 scores of \textsc{Gemma3 12B}, \textsc{Qwen3 14B}, and \textsc{Exaone 7.8B} under different conflict levels.}
\label{tab:gemma_qwen_exaone_em_f1_conflict}
\end{table*}

\section{Benchmark Evaluation}
We conduct a series of experiments to evaluate multilingual LLM performance on our locale-aware QA dataset. Our goal is to assess how well current models handle both locale-invariant and locale-sensitive questions, and to quantify the limitations of existing evaluation protocols when applied to culturally or regionally diverse inputs.

\subsection{Experimental Setup}
We evaluate five widely used large language models with multilingual capabilities: gpt4-1, Qwen 3, Gemma 3, LLaMA 3.1 and EXAONE. These models vary in architecture, size, and pretraining corpora, representing a broad range of capabilities in multilingual understanding and generation. 

All models are evaluated in a zero-shot QA setting without fine-tuning. For each QA pair, the model generates an answer using a consistent prompting format adapted for the language. We apply two evaluation metrics:

\begin{itemize}
    \item \textbf{Exact Match (EM)}: A binary metric that assigns 1 if the predicted answer exactly matches any of the reference answers, and 0 otherwise:
    \[
    \text{EM} = 
    \begin{cases}
    1, & \text{if } \text{prediction} = \text{reference} \\
    0, & \text{otherwise}
    \end{cases}
    \]
    
    \item \textbf{F1 Score}: Measures the token-level overlap between the predicted and reference answers. It is computed as the harmonic mean of precision and recall:
\textbf{F1 Score} measures the token-level overlap between the prediction and the reference answer. It is computed as the harmonic mean of precision and recall:
\begin{equation}
\text{Precision} = \frac{|\text{Prediction} \cap \text{Reference}|}{|\text{Prediction}|}
\end{equation}

\begin{equation}
\text{Recall} = \frac{|\text{Prediction} \cap \text{Reference}|}{|\text{Reference}|}
\end{equation}

\begin{equation}
\text{F1} = \frac{2 \cdot \text{Precision} \cdot \text{Recall}}{\text{Precision} + \text{Recall}}
\end{equation}
\end{itemize}

We evaluate both locale-invariant and locale-aware settings.

\subsection{Main Results}

\paragraph*{(1) Performance gap between English and other languages.}

Table~\ref{tab: main} presents the performance of five LLMs on the XLQA benchmark. While English achieves the highest scores across all models, performance on other languages drops, particularly for those involving culturally diverse or underrepresented regions such as Arabic, Hebrew, Korean, and Vietnamese. This suggests that despite multilingual pretraining, current models struggle to generalize locale-aware reasoning beyond high-resource languages like English.
\paragraph*{(2) Performance degradation on culturally sensitive questions.}  
Table~\ref{tab:gemma_qwen_exaone_em_f1_conflict} offers a more granular view by separating questions into \textit{non-conflict} and \textit{least-conflict} subsets. The results show a consistent and substantial performance drop across all models when faced with locale-sensitive questions. This highlights that answering such questions effectively requires not only understanding the language but also retaining culturally grounded knowledge specific to each region.
Interestingly, models trained with a regional focus tend to perform better on conflict questions in their respective languages. For example, \textsc{Exaone} achieves the highest conflict F1 score on Korean and \textsc{Qwen3} on Chinese. While exact language-wise pretraining proportions are not publicly disclosed, these results suggest that higher exposure to specific locale-language data during pretraining enables models to better handle culturally nuanced inputs in that region.

\begin{table*}[h]
\centering
\resizebox{\textwidth}{!}{%
\begin{tabular}{lp{0.25\linewidth}p{0.5\linewidth}r}
\toprule
\textbf{Conflict Type} & \textbf{Subtopics (Categories)} & \textbf{Representative Questions} & \textbf{Conflict Count} \\
\midrule
\textbf{Entity Conflict} & 
Music, TV actors, Sports players & 
\textit{who sang oh what a night?}, \textit{who played TJ on Head of the Class?}, \textit{who is the coach for the Toronto Raptors?} & 1032 \\
\textbf{Factual Conflict} & 
Geography, Political history, Team records & 
\textit{how many states does the rocky mountains cover?}, \textit{when was the last time the Lakers made the playoffs?} & 431 \\
\textbf{Cultural Reference} & 
TV show winners, Music awards, Famous media & 
\textit{who won America's Got Talent in 2015?}, \textit{who has the most Grammys?} & 512 \\
\textbf{Ambiguous Question} & 
Religion, Social media, General trivia & 
\textit{who wrote the Book of Lamentations?}, \textit{who has the most Instagram followers?} & 381 \\
\bottomrule
\end{tabular}%
}
\caption{Conflict-inducing questions categorized by conflict type, with subtopics and representative examples.}
\label{tab:conflict_analysis}
\end{table*}

\subsection{Prompt Sensitivity}

We examine the impact of prompt design using Qwen3 across four variants: \textbf{EN} (English prompt) and \textbf{EN-LOC} (English with locale emphasis).

Table~\ref{prompt} shows that prompts with explicit locale guidance (EN-LOC) improve accuracy, especially for culturally sensitive languages like Arabic and Korean. However, over-conditioning can sometimes lead to stereotype-driven outputs. 
While EN-LOC prompts generally improve performance, the degree of improvement varies significantly across languages. The gains are especially pronounced in Japanese (+25.03 F1), Chinese (+17.42), and Korean (+7.58), suggesting that locale-specific grounding is particularly beneficial in languages with strong locale reference frames.

\begin{table}[ht]
\centering
\label{tab:qwen_f1_lang_overall}
\begin{tabular}{lcc}
\toprule
Lang & EN & EN-LOC \\
\midrule
en       & 48.03 & 49.43 \\
ko       & 12.33 & 19.91 \\
ar       & 11.93 & 19.30 \\
he       & 16.37 & 16.08 \\
ja       & 19.00 & 44.03 \\
ru       & 16.41 & 27.97 \\
vi       & 33.37 & 39.34 \\ 
zh\_cn   & 30.15 & 47.57 \\ \bottomrule
\textbf{Overall} & 23.45 & 32.95 \\
\bottomrule
\end{tabular}
\caption{Performance across languages under different prompting strategies on Qwen3.}
\label{prompt}
\end{table}


\begin{figure}[h]
    \centering
    \includegraphics[width=0.9\linewidth]{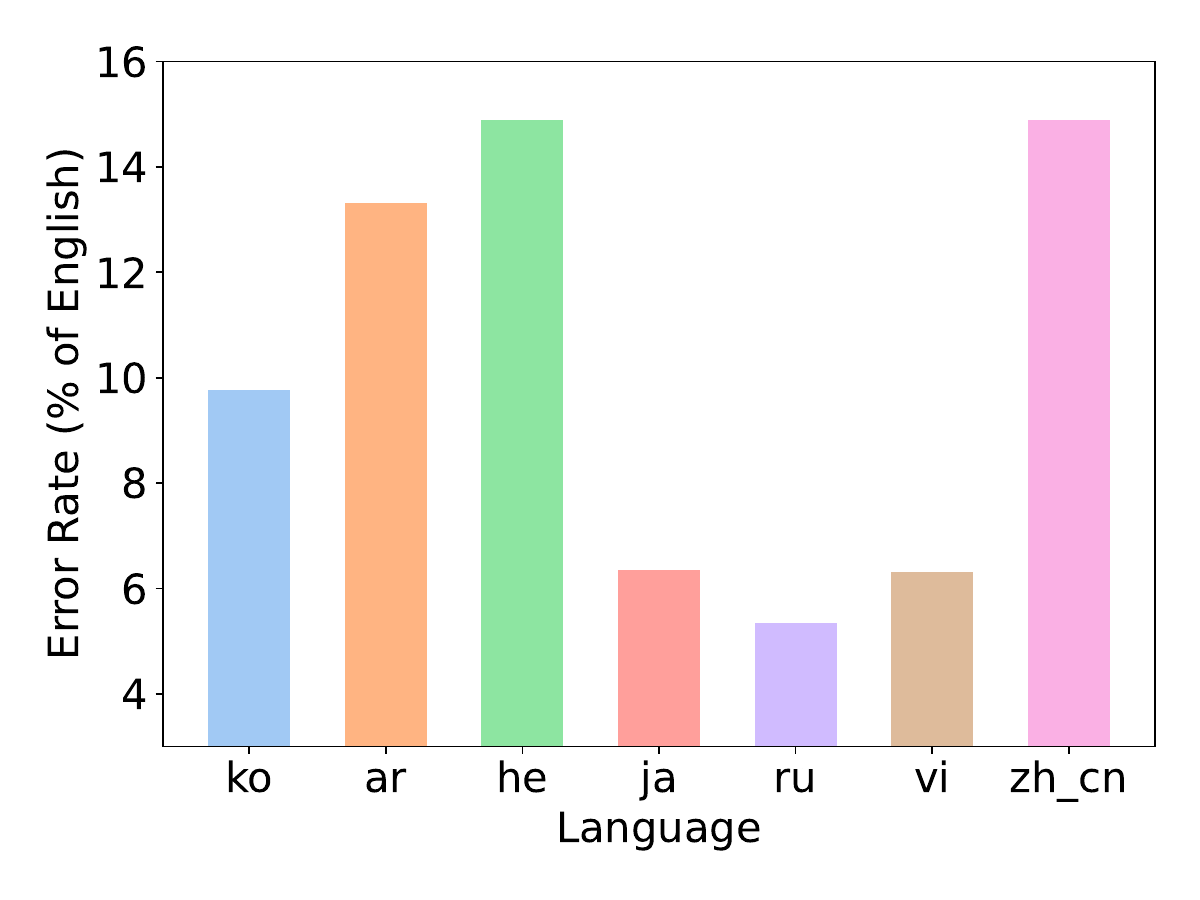}
    \caption{Comparison of translation error rates between naive translation and our back-translation pipeline.}
    \label{fig:error_rate_comparison}
\end{figure}

\begin{table}[ht]
\resizebox{1.\columnwidth}{!}{%
\begin{tabular}{lcc}
\toprule
\textbf{Lang}                                    & \textbf{Conflicted Answers}             & \textbf{Conflict Rate (\%)}    \\ \midrule
ar                                          & 1471                                     & 46.2\%                         \\
he                                          & 1413                                     & 44.3\%                         \\
ja                                          & 1044                                     & 32.8\%                         \\
ru                                          & 963                                      & 30.2\%                         \\
ko                                          & 1188                                     & 37.3\%                         \\
zh\_cn                                      & 1242                                     & 39.0\%                         \\
vi                                          & 909                                      & 28.5\%                         \\ \midrule
At Least One Conflict                                & 2356                                     & 73.9\%                         \\ \bottomrule
\end{tabular}
} %
\caption{Language-wise distribution of answer conflicts in the XLQA benchmark.}
\label{tab:consistency}
\end{table}

\subsection{Ensuring Semantic Consistency in Multilingual Questions}
A back-translation-based filtering helps identify and remove mistranslations that may introduce unintended meaning shifts during naive machine translation. As shown in Figure~\ref{fig:error_rate_comparison}, our back-translation pipeline significantly reduces translation error rates across most languages, particularly in Arabic, Hebrew, and Chinese languages that often exhibit greater semantic divergence from English. By improving the alignment between original and translated questions, this filtering step enhances the overall quality and reliability of locale-sensitive evaluation.

\subsection{Categorization of Conflict-Inducing Questions}
To better understand the sources of semantic divergence across languages, we manually categorize a subset of conflict-inducing questions based on the nature of the discrepancy observed in answers. This typology enables a more fine-grained analysis of the types of ambiguity and regional variability that arise in multilingual QA.

We categorize conflict-inducing questions into four types. These include \textit{Entity Conflict}, \textit{Factual Conflict}, \textit{Cultural Reference}, and \textit{Ambiguous Question}. \textbf{Entity Conflict} refers to cases where the referent entity varies across locales due to differing popularity or interpretation, such as entertainers or sports figures. \textbf{Factual Conflict} includes questions grounded in historical or statistical facts that may be represented differently depending on regional data sources. \textbf{Cultural Reference} covers instances involving awards, media, or events where local recognition or framing differs. Finally, \textbf{Ambiguous Question} includes vague or broadly interpretable queries that elicit culturally biased or interpretive responses.

Table~\ref{tab:conflict_analysis} summarizes each conflict type along with representative subtopics, example questions, and the number of instances observed in our annotated subset. Entity-related conflicts were the most frequent, accounting for 1,032 questions, followed by Cultural References and Factual Conflicts. This distribution highlights the significant role of culturally grounded knowledge and localized salience in generating cross-lingual answer variability.

\begin{table}[t]
\centering
\resizebox{\columnwidth}{!}{%
\begin{tabular}{lcccccccc}
\toprule
\textbf{} & \textbf{en} & \textbf{ar} & \textbf{he} & \textbf{ja} & \textbf{ko} & \textbf{ru} & \textbf{zh\_cn} & \textbf{vi} \\
\midrule
\textbf{Avg Question Length} & 37 & 33 & 31 & 26 & 22 & 40 & 17 & 38 \\
\textbf{Avg Answer Length}   & 5  & 5  & 5  & 8  & 4  & 5  & 6  & 5 \\
\bottomrule
\end{tabular}%
}
\caption{Average question and answer lengths across languages (rounded to nearest integer).}
\label{tab:avg_lengths}
\end{table}

\section{Conclusion}

In this work, we identify a critical gap in existing multilingual QA benchmarks, the lack of consideration for locale-specific knowledge and culturally valid answer divergence. While prior evaluations assume semantic equivalence and a single correct answer across languages, our analysis shows that this assumption fails in questions involving cultural or regional context. To address this, we propose a method for constructing locale-aware evaluation subsets that allow for valid answer variation across languages. Our approach combines translation consistency checks and prompt-based answer divergence detection to identify culturally sensitive questions. We demonstrate that such questions are not rare, and that standard evaluation protocols may underestimate the capabilities of multilingual models in diverse linguistic settings. This work calls for a shift in multilingual QA evaluation toward frameworks that are not only linguistically fair but also culturally grounded.

\section*{Limitations}


Our evaluation may be inherently bounded by the capabilities of the proprietary large language models (LLMs) accessed via API. Since these models serve as oracle systems for translation and answer generation, their performance imposes an upper bound on the quality and diversity of our data. To mitigate potential issues arising from translation artifacts or inconsistencies, we applied a semantic consistency filtering step using back-translation and LLM-as-judge comparison to ensure that the generated multilingual questions preserve the meaning of the original seed questions. Additionally, due to computational resource constraints, we were unable to include larger-scale open-source multilingual models that require substantial local infrastructure. To compensate for this limitation, we evaluated a diverse set of models—both proprietary and open-source—covering a range of capabilities and linguistic domains, and conducted all evaluations under a unified framework to ensure comparability. Future work could expand this line of research by integrating scalable open-source multilingual models in controlled environments and broadening the linguistic and regional scope of the evaluation.

\section{Acknowledgment}
This work was supported by the Agency For Defense Development by the Korean Government(UI247035TF)

\bibliography{latex/custom}

\appendix
\section{XLQA Construction Details}

\subsection{Prompt Templates}
We provide the full prompt templates used throughout the XLQA benchmark construction and evaluation pipeline. These include:

\textbf{Translation prompts}, used to generate multilingual versions of questions from English.
\begin{mdframed}[backgroundcolor=gray!10, linewidth=0.5pt]

Given the question: \{question\}, please translate it into \{loc\}. Just output the translated question only, with no comments or formatting.
\end{mdframed}

\vspace{5pt}

\textbf{Back-translation prompts}, used to back-translate to English.
\begin{mdframed}[backgroundcolor=gray!10, linewidth=0.5pt]
\raggedright
Given the question: \{translated\_response.output\_text\}, please translate it back into English. Just output the translated question only, with no comments or formatting.
\end{mdframed}

\vspace{5pt}

\textbf{Consistency filtering prompts}, used to verify semantic consistency across languages.
\begin{mdframed}[backgroundcolor=gray!10, linewidth=0.5pt]
\raggedright
Given the question: \{question\}, please check if the back translation: \{back\_translation.output\_text\} is correct.  
If it is correct, output "yes". If it is not correct, output "no".
\end{mdframed}

\vspace{5pt}

\textbf{Locale-aware answer generation prompts}, which condition the model to generate region-specific answers if appropriate.
\begin{mdframed}[backgroundcolor=gray!10, linewidth=0.5pt]
\raggedright
You are given a question and its answer in a specific language.  
If the answer is a full sentence or unnecessarily long, rewrite it as a short, direct answer — a concise phrase or a single word — while preserving the original language.  
If the answer is already short and direct, leave it unchanged.  
Do not add any explanation, translation, or formatting.

Input:  
Question: \{question\}  
Answer: \{answer\}

Output:  
Only the final, concise answer in the same language as the input.
\end{mdframed}

\subsection{Human Verification Agreements Ratio}
\begin{table*}[h]
\centering
\resizebox{\textwidth}{!}{%
\begin{tabular}{lcccc}
\toprule
\textbf{Language} & \textbf{Correctness (3/3)} & \textbf{Correctness ($\geq$2/3)} & \textbf{Sensitivity (3/3)} & \textbf{Sensitivity ($\geq$2/3)} \\
\midrule
English (en)    & 91.2\% & 98.5\% & 88.3\% & 96.7\% \\
Korean (ko)     & 89.7\% & 97.4\% & 85.2\% & 95.9\% \\
Arabic (ar)     & 86.4\% & 96.1\% & 80.5\% & 93.8\% \\
Hebrew (he)     & 88.1\% & 97.0\% & 82.7\% & 94.6\% \\
Japanese (ja)   & 90.5\% & 98.1\% & 87.0\% & 96.2\% \\
Russian (ru)    & 87.9\% & 96.8\% & 84.1\% & 94.3\% \\
Vietnamese (vi) & 89.3\% & 97.9\% & 86.5\% & 95.7\% \\
Chinese (zh\_cn)& 88.7\% & 97.5\% & 83.6\% & 94.8\% \\
\midrule
\textbf{Average} & \textbf{88.9\%} & \textbf{97.4\%} & \textbf{84.7\%} & \textbf{95.3\%} \\
\bottomrule
\end{tabular}%
}
\caption{Annotator agreement rates by language. The table shows the percentage of instances where all three annotators (3/3) or at least two annotators (2/3) agreed on correctness and locale-sensitivity labels.}
\label{tab:annotation_agreement}

\end{table*}

\subsection{Locale Sensitivity Annotation Guidelines}
We define a question as \textit{locale-sensitive} if its correct answer may differ depending on regional, cultural, or national context, even when the semantic intent of the question remains the same.

Annotators were instructed to mark a question as locale-sensitive if:

Regionally salient knowledge affects the expected answer (e.g., “most famous tower”).

Political, institutional, or cultural prominence varies by country or language group.

The question involves subjective norms or identity references (e.g., “national dish”, “popular leader”).

Borderline cases were resolved by majority voting across annotators with multilingual and regional backgrounds.

\section{Experimental Details}
\subsection{Models}
We use the following models in our experiments:

\begin{itemize}    
    \item \textbf{Gemma3 12B}: Uses Gemma3 with 12B parameters. Licensed under \textbf{Apache 2.0 license.}.
    
    \item \textbf{Qwen3 14B}: Uses Qwen3 with 14B parameters. Licensed under the \textbf{Apache 2.0 license.}.
    
    \item \textbf{LLaMA-3.1 8B}: Has 8B parameters and is released under the \textbf{LLaMA 3 Community License Agreement}.
    
    \item \textbf{GPT-4.1}: These models are not open-source and are accessible only via API requests. They are governed by \textbf{proprietary licenses}.
    
    \item \textbf{Exaone 7.8B}: Uses Exaone with 7.8B parameters. Licensed under \textbf{EXAONE AI Model License Agreement}.
\end{itemize}

All the models set the temperature to 0.

\subsection{Budget}
We use the RTX A6000 GPU X 1 with 20 hours. 

\section{Human Annotation}

\begin{figure*}
    \centering
    \includegraphics[width=\textwidth]{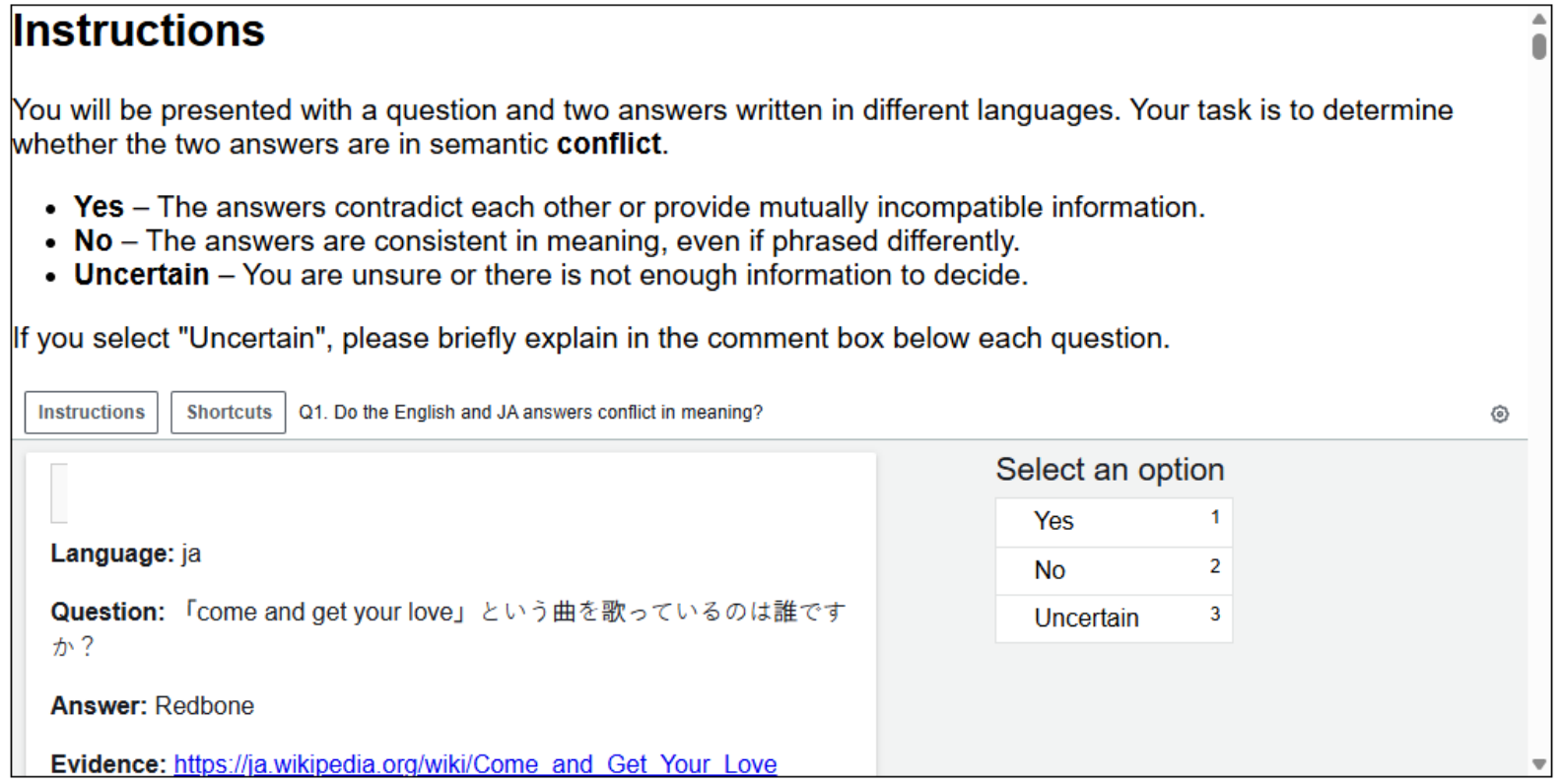}
    \caption{
        \textbf{Survey screenshot.} Interface shown to MTurk annotators during the human verification stage.
    }
    \label{fig:mturk}
\end{figure*}

To verify the correctness and locale sensitivity of the model-generated answers, we conducted human annotation using Amazon Mechanical Turk (MTurk). For each language, we recruited \textbf{three independent annotators} who are speakers of the respective target language to evaluate each QA-evidence triple. Annotators were presented with the original question, the model-generated answer, and its associated supporting evidence (e.g., URL or passage), and were instructed to assess as in Figure~\ref{fig:mturk}.

`

Each annotation instance was reviewed by three annotators. Final labels were determined via \textbf{majority voting}. Annotator agreement rates are summarized in Table~\ref{tab:annotation_agreement}.

All annotators were compensated fairly according to MTurk standards, and informed that their responses would be used for research purposes. No personally identifiable information was collected during the process. Tasks involving potentially sensitive content were manually reviewed and filtered prior to annotation to avoid harm or discomfort.

\section{Ethical Considerations}
While XLQA promotes cultural inclusion in QA evaluation, locale-aware generation introduces ethical challenges. Prompts conditioned on locale risk overgeneralization or reinforcement of cultural stereotypes. We manually reviewed outputs for offensiveness and excluded instances containing bias or politically sensitive content.

Furthermore, hallucination in low-resource languages may amplify misinformation if locale grounding is weak. We recommend that future work incorporate human validation when deploying such systems in high-stakes settings.

\label{sec:appendix}

\end{document}